\documentclass[10pt,twocolumn,letterpaper]{article}

\usepackage{cvpr}              
\definecolor{cvprblue}{rgb}{0.21,0.49,0.74}
\usepackage[pagebackref,breaklinks,colorlinks,allcolors=cvprblue]{hyperref}
\usepackage{booktabs}
\usepackage[table]{xcolor}
\usepackage{xcolor}
\usepackage{multirow}
\usepackage{graphicx}
\usepackage{pifont} 

\definecolor{darkgreen}{RGB}{0,120,0}  
\newcommand{\cmark}{\textcolor{darkgreen}{\ding{51}}} 
\newcommand{\xmark}{\textcolor{red}{\ding{55}}}   

\def\paperID{12222} 
\def\confName{CVPR}
\def\confYear{2026}

\title{GUI-CEval: A Hierarchical and Comprehensive Chinese Benchmark\\ for Mobile GUI Agents}

\author{%
  Yang Li$^{1}$\thanks{These authors contributed equally.}, Yuchen Liu$^{1*}$, Haoyu Lu$^{1}$, Zhiqiang Xia$^{1}$, Hongzhen Wang$^{1}$, Kaiyang Han$^{1}$, \\Changpeng Yang$^{1}$, Jinyang Wu$^{1}$, Jiaming Xu$^{1}$, Runyu Shi$^{1}$\thanks{Corresponding author}, Ying Huang$^{1}$ \\
  $^1$ HyperAI Team, Xiaomi Corporation \\
  \{liyang134,xujiaming1\}@xiaomi.com
  }

\begin{document}

\maketitle

\begin{abstract}


Recent progress in Multimodal Large Language Models (MLLMs) has enabled mobile GUI agents capable of visual perception, cross-modal reasoning, and interactive control. However, existing benchmarks are largely English-centric and fail to capture the linguistic and interaction characteristics of the Chinese mobile ecosystem. They also focus on isolated skills such as GUI grounding or offline agent, lacking a unified and fine-grained framework to assess the full capability chain from perception to execution. To address this gap, we introduce \textbf{GUI-CEval}, the first comprehensive benchmark for Chinese mobile GUI agents, built entirely on \textbf{physical device} environments. GUI-CEval spans 201 mainstream apps across four device types and adopts a two-level structure that evaluates both atomic abilities and realistic application-level performance along five dimensions: perception, planning, reflection, execution, and evaluation. All data are collected and verified through multi-stage manual processes to ensure authenticity and reproducibility. Extensive experiments on 20 representative MLLMs and multi-agent systems show that while models such as Qwen2.5-VL and UI-TARS perform competitively, most MLLMs still exhibit clear weaknesses in reflective decision-making and post-action self-evaluation, limiting their reliability in real-world interactions. We hope GUI-CEval provides a comprehensive  and interpretable benchmark to guide capability diagnosis and advance the development of Chinese mobile GUI agents.

\end{abstract}

\begin{figure}[t]
    \centering
    \includegraphics[width=0.95\linewidth]{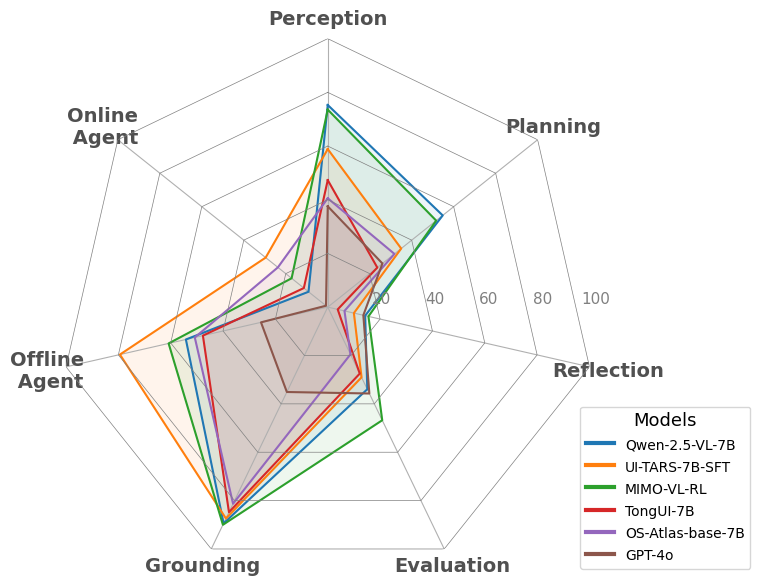}
    \caption{Results of six representative multimodal large language models across seven tasks defined in GUI-CEval. The uneven radar profiles show that GUI-CEval offers a comprehensive examination of perception–to–execution capabilities and poses a substantial challenge to current models.
}
    \label{ratar}
\end{figure}    
\section{Introduction}
\label{sec:intro}




\begin{figure*}[!t]
   \centering
   \includegraphics[width=\linewidth]{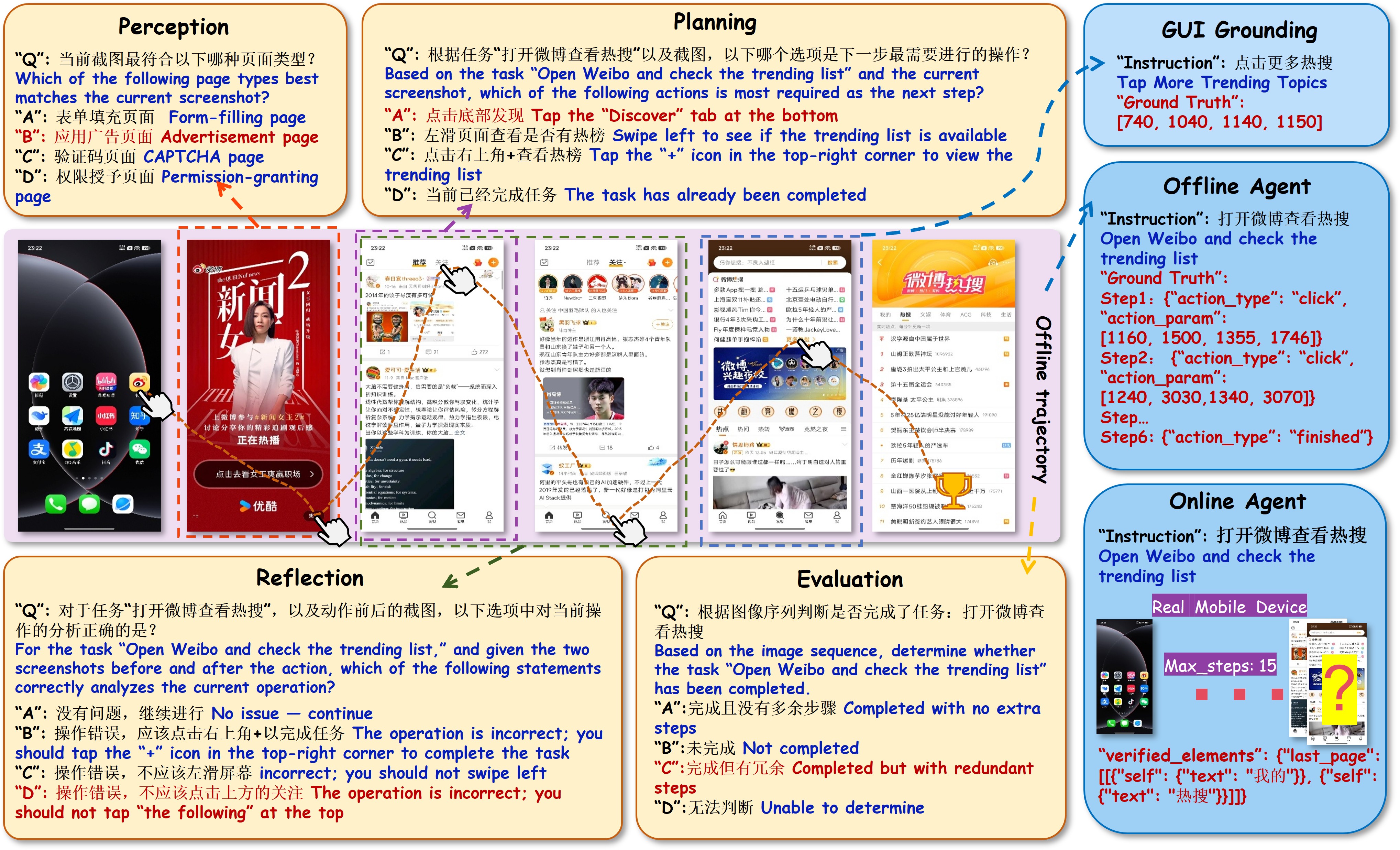}
   \caption{A representative example illustrating how GUI-CEval provides a comprehensive analysis of a Chinese mobile instruction. It evaluates realistic GUI application ability using tasks constructed from real mobile environments and trajectories, while also assessing atomic skills via single-answer multiple-choice questions, enabling developers to diagnose and improve model weaknesses.}
   \label{example}
\end{figure*}



The rapid advancement of multimodal large language models (MLLMs) \cite{bai2025qwen2,zhu2025internvl3,coreteam2025mimovltechnicalreport,yao2024minicpm} has empowered GUI Agents with the ability to perceive, reason, and act within real graphical interfaces \cite{shen2025mind,zheng2025pptagent}, enabling intelligent interaction and automation across mobile environments \cite{liu2024autoglm,wang2025ui,ye2025mobile}.


Although many existing benchmarks such as Screenspot \cite{cheng2024seeclick}, Screenspot Pro \cite{li2025screenspot}, AndroidControl \cite{li2024effects}, and AndroidWorld \cite{rawlesandroidworld} have advanced the capabilities of GUI Agents, several limitations persist: (i) \textbf{language bias}: most are English-centric \cite{liuvisualwebbench,deng2024mobile,liu2024visualagentbench}, limiting evaluation in Chinese ecosystems \cite{zhang2025agentcpm}; (ii) \textbf{scene inconsistency}: data are collected from diverse platforms \cite{cheng2024seeclick,wang2025mmbench}, lacking focused assessment on mobile environments; (iii) \textbf{task narrowness}: current benchmarks emphasize UI element localization \cite{cheng2024seeclick,li2025screenspot} or offline agent success rates \cite{li2024effects,chai2025amex}, offering limited insight into comprehensive assessment and full-pipeline capabilities; and (iv) \textbf{data realism}: automated collection \cite{li2025autogui,shi2017world}  and validation overlook \textit{real user intents}, reducing practical reliability.


To establish a fair, comparable, and diagnostic evaluation standard for Chinese mobile environments, we propose GUI-CEval, the first comprehensive benchmark for mobile GUI agents tailored to the Chinese ecosystem, as shown in Fig.~\ref{example}. GUI-CEval spans 201 mainstream Chinese applications across four real mobile device types, consisting of 4,028 agent tasks and 4,194 multimodal question–answering (QA) tasks. The benchmark adopts a hierarchical design that integrates both fundamental and applied capabilities, defining five core dimensions aligned with the complete workflow of a mobile GUI agent: perception, planning, reflection, execution, and evaluation. The fundamental capabilities are examined through diagnostic multimodal QA tasks, focusing on atomic skills, enabling fine-grained capability analysis, and guiding model improvement. While the application tasks cover three critical scenarios—GUI grounding, offline agent, and online agent—to assess end-to-end performance from target localization to action execution. All data are human-curated on real mobile devices and verified, ensuring realistic interaction contexts and coherent task flows \cite{huang2023c}. This design substantially improves the credibility, realism, and diagnostic reliability of GUI agent evaluation. The datasets and evaluation code will be released to advance the development of Chinese mobile GUI agents.

We evaluate 20 representative multimodal models, GUI-specific models, and multi-agent systems on GUI-CEval. Results in Fig.~\ref{ratar} reveal that current models are not yet ready for stable deployment in real Chinese mobile environments. In summary, our contributions are as follows:
\begin{enumerate}
\item \textbf{Comprehensive Chinese mobile benchmark.}
We present \textbf{GUI-CEval}, the first large-scale benchmark for Chinese mobile GUI agents, covering 201 mainstream apps and 4 real device types with 4,194 multimodal QA and 4,028 agent tasks for comprehensive and fine-grained evaluation.

\item \textbf{Hierarchical five-dimensional diagnostic framework.}  
GUI-CEval introduces a hierarchical structure spanning \textit{perception}, \textit{planning}, \textit{reflection}, \textit{execution}, and \textit{evaluation}. It unifies GUI grounding, offline, and online agent scenarios to enable fine-grained, end-to-end capability diagnosis.

\item \textbf{Human-verified real-world data pipeline.}  
All data are collected and annotated through real device demonstrations and human review, ensuring realistic interaction contexts and preventing data leakage or template bias.

\item \textbf{Extensive evaluation and insights.}  
Experiments on 20 representative models reveal persistent weaknesses in generalization, stability, and reflective reasoning, highlighting GUI-CEval’s value as a diagnostic and developmental foundation for Chinese mobile GUI agents.

\end{enumerate}
\section{Related Work}

\paragraph{GUI Agent.}
Current approaches generally fall into two main paradigms:
(i) The workflow paradigm \cite{wang2024mobile,wang2024mobilev2}, which uses carefully designed prompts containing task descriptions, UI states, action histories, and reflective reasoning, often combined with grounding \cite{liu2024grounding} or OCR \cite{poznanski2025olmocr} models to form multi-component execution pipelines;
(ii) The end-to-end paradigm \cite{cheng2024seeclick,xuaguvis,lu2025ui,luo2025gui,tang2025gui}, which trains MLLMs through supervised learning  or reinforcement learning to unify perception, reasoning, and action generation within a single model, offering better execution speed and lower resource consumption. 
Despite these advances, mobile GUI agents still rely heavily on robust error recovery mechanisms and fall far short of human performance in real-world scenarios.

\paragraph{GUI Benchmarks.} Existing GUI benchmarks mainly fall into three categories: (i) Grounding benchmarks, such as Screenspot \cite{cheng2024seeclick}, Screenspot Pro \cite{li2025screenspot}, UI-E2I \cite{liu2025ui}, CAGUI \cite{zhang2025agentcpm}, and UI-Vision \cite{nayakui}, that operate on static screenshots to test alignment between natural-language instructions and interface region; (ii) Offline Agent benchmarks, such as AndroidControl \cite{li2024effects}, AITW \cite{rawles2023androidinthewild}, and Mind2Web \cite{deng2023mind2web}, that evaluate action prediction from recorded trajectories; (iii) Online Agent benchmarks, such as AndroidWorld \cite{rawlesandroidworld}, OSWorld \cite{xie2024osworld}, and SPA-Bench \cite{chen2024spa}, that assess full execution on real or simulated devices. A few benchmarks focus on foundation-level capabilities \cite{wang2025mmbench,liuvisualwebbench}, evaluating holistic layout and functional understanding via multimodal QA, as in MMBench-GUI \cite{wang2025mmbench}. 

However, current resources are mostly English-only, uneven across platforms, and fragmented by task type, as shown in Tab.~\ref{related_tab}. They often cover only one stage of perception or control, making it hard to pinpoint failure causes such as grounding errors, planning issues, or domain shift. Cross-benchmark metrics (e.g., success rate, step count) are also not comparable due to divergent task complexity. Meanwhile, Chinese mobile interfaces remain largely underrepresented. These gaps highlight the need for a unified evaluation framework that spans the full interaction chain, supports fine-grained diagnosis, and enables fairer comparison across languages and platforms.

\begin{table}[t]
\centering
\caption{Comparison of existing GUI evaluation datasets with our GUI-CEval. \textbf{F}, \textbf{G}, \textbf{Off}, \textbf{On} denote the tasks of \textbf{F}oundation, \textbf{G}rounding, \textbf{Off}line Agent and \textbf{On}line Agent.}
\resizebox{\linewidth}{!}{
\begin{tabular}{l|cccccccc}
\toprule
\textbf{Dataset} & \textbf{Language} & \textbf{Number} & \textbf{Platform} &
\textbf{F} & \textbf{G} & \textbf{Off} & \textbf{On} \\
\midrule
Screenspot \cite{cheng2024seeclick}        & EN     & 1272  & Mobile, Desktop, Web & 
\xmark & \cmark & \xmark & \xmark \\
Screenspot Pro \cite{li2025screenspot}    & EN+CN  & 1581  & Desktop               & 
\xmark & \cmark & \xmark & \xmark \\
MMbench-GUI \cite{wang2025mmbench}      & EN     & 8123  & Mobile, Desktop, Web  & 
\cmark & \cmark & \cmark & \xmark \\
UI-E2I \cite{liu2025ui}           & EN     & 1477  & Desktop               & 
\xmark & \cmark & \xmark & \xmark \\
CAGUI  \cite{zhang2025agentcpm}           & CN     & 3600  & Mobile                &
\xmark & \cmark & \cmark & \xmark \\
UI-Vision \cite{nayakui}        & EN     & 8227  & Desktop               &
\xmark & \cmark & \cmark & \xmark \\
AndroidControl \cite{li2024effects}  & EN     & 1412  & Mobile                &
\xmark & \xmark & \cmark & \xmark \\
AITW   \cite{rawles2023androidinthewild}           & EN     & 30378 & Mobile                &
\xmark & \xmark & \cmark & \xmark \\
Mind2Web \cite{deng2023mind2web}         & EN     & 2350  & Web                   &
\xmark & \xmark & \cmark & \xmark \\
VisualWebBench \cite{liuvisualwebbench}   & EN     & 1500  & Web                   &
\cmark & \xmark & \xmark & \xmark \\
OSWorld  \cite{xie2024osworld}         & EN     & 369   & Desktop               &
\xmark & \xmark & \xmark & \cmark \\
AndroidWorld \cite{rawlesandroidworld}    & EN     & 116   & Mobile                &
\xmark & \xmark & \xmark & \cmark \\
SPA-BENCH  \cite{chen2024spa}       & EN+CN  & 340   & Mobile                &
\xmark & \xmark & \xmark & \cmark \\
\midrule
\rowcolor{gray!20}
\textbf{GUI-CEval} & \textbf{CN} & \textbf{8222} & \textbf{Mobile} &
\textbf{\cmark} & \textbf{\cmark} & \textbf{\cmark} & \textbf{\cmark} \\
\bottomrule
\end{tabular}}

\label{related_tab}
\end{table}
\section{GUI-CEval}

GUI-CEval aims to establish a fair, comparable, and diagnostic evaluation standard for Chinese mobile environments, aligned with the long-term vision of a “super personal device agent.” It jointly emphasizes reliable execution—selecting the correct targets and actions—and systematic assessment of dialog-level understanding and chain-level decision making. 
The benchmark adopts a two-tier structure. The foundation task decomposes atomic skills through multimodal QA and provides fine-grained diagnostics on five dimensions: perception, planning, reflection, execution, and evaluation. The application task unifies GUI grounding, offline agents, and online agents within the same application domain and page family on real mobile devices.

\begin{figure*}[!t]
   \centering
   \includegraphics[width=0.96\linewidth]{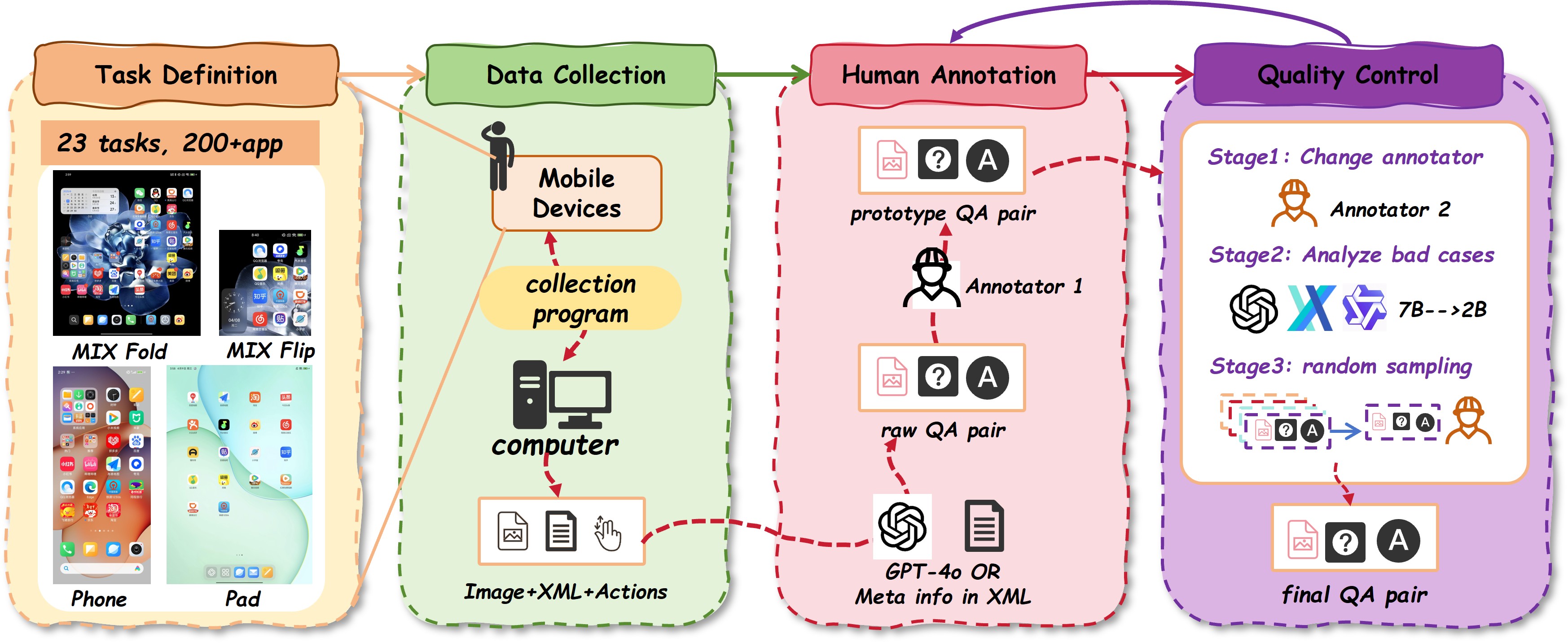}
   \caption{The pipeline of GUI-CEval, covering task definition, data collection, human annotation, and multi-stage quality control. }
   \label{fig:framework}
\end{figure*}

\subsection{Task Definition}

\subsubsection{Foundation Tasks}

Foundation tasks evaluate core model capabilities in mobile GUI settings in a unified, comparable, and diagnostic way, while avoiding the prompt sensitivity and environmental noise of full end-to-end interaction. The task design follows three principles: (i) assess grounding and decision-making skills through multi-modal QA rather than real interaction, enabling stable scoring that separates expression errors from understanding errors; (ii) decouple atomic skills along the execution pipeline to improve attribution and provide interpretable links to application-level behavior; (iii) include process-and outcome-level judgment tasks to support reward modeling \cite{sun2025genesis} and self-reflection. These designs cover capability areas overlooked by prior benchmarks and yield a more reliable, actionable characterization of model strengths and weaknesses in real mobile scenarios. All tasks adopt a unified prompt format requiring a single choice, with accuracy as the primary metric. Detailed task descriptions and statistics are provided in Appendix A.2.


\paragraph{Perception Task.} Accurate recognition of the current app, page, actionable widgets, and on-screen text is essential for downstream planning and execution, yet prior benchmarks rarely examine these skills in depth. GUI-CEval therefore decomposes element grounding into appearance, spatial, and functional cues to separately assess visual representation, geometric reasoning, and semantic alignment, and adds a reverse task that generates natural-language descriptions from specified regions to test perception–expression consistency. To reduce sensitivity to coordinate phrasing, we use Set-of-Marks (SoM) candidate boxes \cite{yang2023set} and require the model to choose from four regions. At the page level, we evaluate app identification and special-page recognition (e.g., payment, ads, permission prompts). At the widget level, we assess screen text understanding and operability recognition to measure comprehension of textual content and interaction affordances.


\paragraph{Planning Tasks.} Given a high-level instruction and the current screenshot, agents must predict the correct next action. We decompose this into task planning and action decision: task planning determines the appropriate next step from the goal and state, testing global understanding and goal decomposition; action decision selects the action type from the feasible space, reflecting judgment over actionable operations. We additionally include action inference to capture causal and state-transition reasoning between consecutive screenshots. All samples in task action inference are normalized to click actions with SoM boxes.


\paragraph{Reflection Tasks.} Reflection measures an agent’s ability to self-check and recover from errors during execution. We design two complementary settings: short-horizon reflection, which evaluates whether a single action is correct given the instruction and nearby frames, and long-horizon reflection, which examines entire trajectories to identify erroneous or redundant steps, capturing cross-step consistency and process-level monitoring.


\paragraph{Evaluation Tasks.} Evaluation focuses on determining task completion and interpreting execution outcomes—crucial for data filtering and reward assignment. We design three complementary tasks: success judgment, which assesses whether the goal is achieved from the instruction and trajectory, ensuring stable reward signals; instruction generation, which infers the most plausible high-level intent from the execution trajectory, testing semantic synthesis; and temporal ordering, which reconstructs the correct step sequence from shuffled screenshots under a given instruction, evaluating causal reasoning and process coherence.

\subsubsection{Application Tasks}


Application evaluation tasks, complementary to foundational capability evaluation, are designed to assess the agent's executability and robustness in real-world environments. We assess GUI grounding, offline agent, and online agent capabilities \textbf{within the same application or homologous pages}. Integrating the evaluation of localization, and decision-making into a unified assessment framework facilitates a more comprehensive evaluation of the model's capabilities and applicable scenarios. Detailed statistics and evaluation are provided in Appendix A.2 and A.4.


\paragraph{GUI Grounding.} Most mobile GUI operations are click-based, and accurate target selection is critical for enabling subsequent actions. Following the ScreenSpot \cite{li2025screenspot} protocol, each task provides a screenshot and a natural language instruction, requiring the model to select the correct interactive location. To reflect real-world usage, targets include text, icons, and images. Performance is measured using point-in-box accuracy, with results stratified by instruction type and difficulty.


\paragraph{Offline Agent.} Offline evaluation replays interactions in static snapshots or prebuilt environments, eliminating system noise to isolate the executability of planning and decision-making. Given a high-level instruction, the current screen, and a short history, the model must iteratively predict the next action and its parameters, such as click coordinates. Offline tasks use the same apps and pages as GUI grounding, aligning localization with execution at the sample level. The primary metric is step-wise accuracy for action type and parameters, supplemented by trajectory-level success rate.


\paragraph{Online Agent.} Online evaluation tests agents on real devices, exposing them to pop-ups, ads, permission prompts, network fluctuations, and timing jitter, thus providing the most realistic measurement. Third-party apps are installed via ADB, and following Android World \cite{rawlesandroidworld} settings, we evaluate common functions that do not require login, while standardizing device specs, resolution, and action space to control environmental variance and ensure reproducibility. Metrics include task success rate, average per-step latency, inference token usage, and total steps. Detailed online agent settings can be found in Appendix A.4.

\begin{figure}[t]
   \centering
   \includegraphics[width=\linewidth]{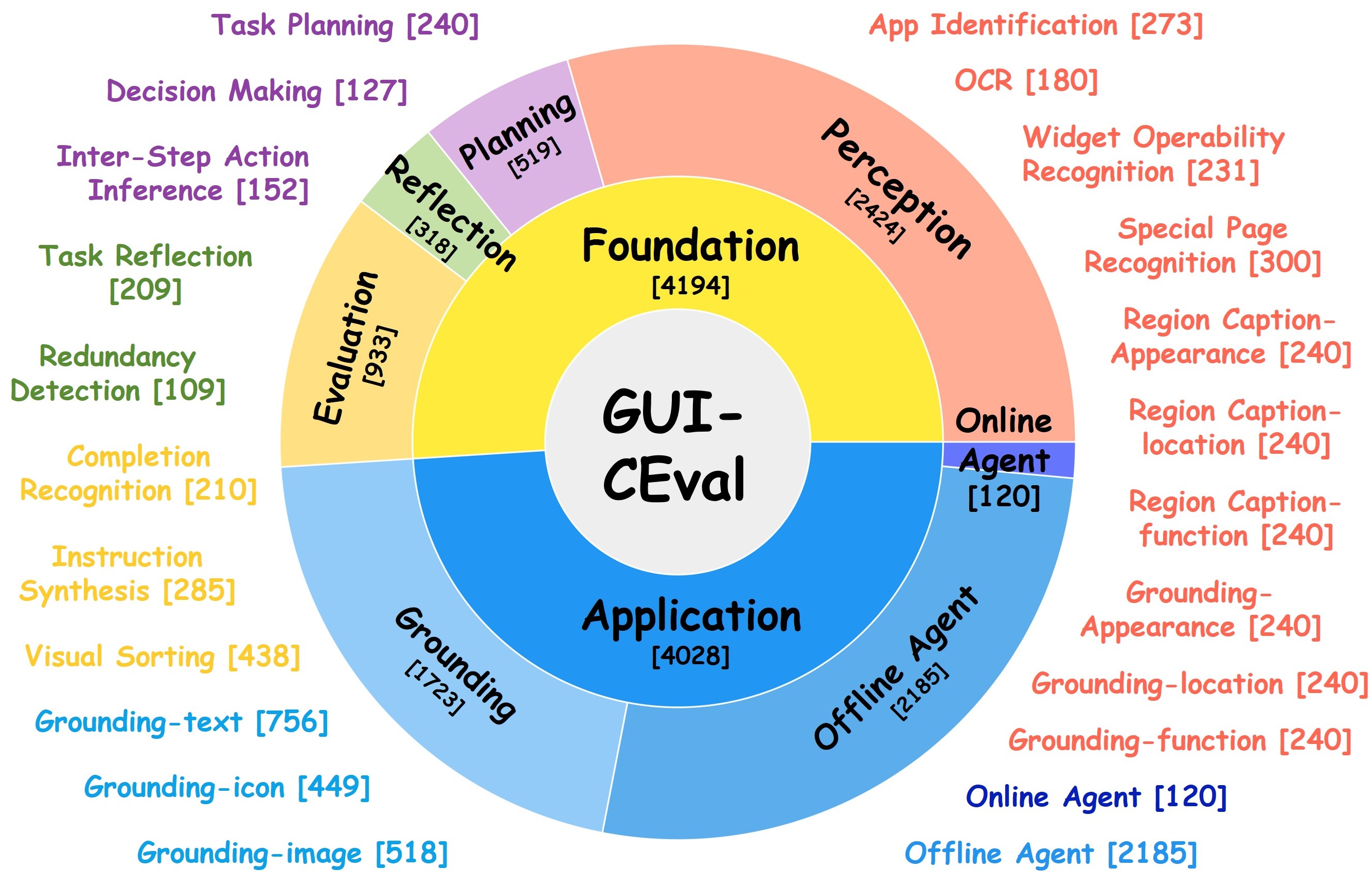}
   \caption{Overall statistics of the GUI-CEval, which consists of 4,194 multimodal QA and 4,028 GUI-agent application tasks.}
   \label{static}
\end{figure}

\subsection{Data Collection and Annotation}


\paragraph{Data Collection.}  The construction process of the dataset is illustrated in Fig.~\ref{fig:framework}. We assembled a device pool spanning various screen sizes and resolutions, including smartphones, tablets, and foldable devices, and used a custom collection tool to capture interactions directly on real devices. Applications were selected from the top 10-20 apps in each category of major app stores, yielding a total of 201 apps that comprehensively reflect daily usage patterns within China’s mobile ecosystem. Each sample includes metadata such as device type, app name, and annotator identity to ensure traceability and verification. 

We collected data through two complementary methods. For single-image collection, annotators captured screenshots and XML files of all major functional pages, including specialized pages such as ads, payment flows, and permission dialogs, while excluding sensitive content. For trajectory collection, 10–20 natural-language instructions were executed on real devices for top 100 high-frequency apps, recording screenshots, XML files, and touch interactions at each step. Only core operations (click, type,  scroll) were included, while automatically triggered events were retained to assess robustness. To reflect realistic error distributions and reduce manual effort, a strong model (e.g. UI-TARS or Qwen2.5-VL) performed predefined tasks, and the resulting trajectories were filtered to remove redundancy and retain failure cases.

\begin{table*}[t]
\centering
\caption{Performance comparison of general-purpose and GUI-specific models on the GUI-CEval benchmark. 
Best results are in \textbf{bold}, and the second-best results are \underline{underlined}. Detailed results for each subtask and all models are provided in Appendix A.4.}
\resizebox{\textwidth}{!}{
\begin{tabular}{l|cccc|ccc|c}
\toprule
Model & Perception & Planning & Reflection & Evaluation & Grounding & Offline Agent & Online Agent & Avg \\
\midrule
\rowcolor{gray!20}
\multicolumn{9}{l}{\textbf{General models}} \\
GPT-4o-mini         & 28.63 & 26.81 &  6.22 & 28.70 &  4.10 & 14.30 &  0.00 & 20.93 \\
GPT-4o              & 37.55 & 26.06 & 13.60 & 35.72 & 35.10 & 25.50 &  0.83 & 27.69 \\
Qwen2.5-VL-3B \cite{bai2025qwen2}       & 65.62 & 42.99 & 12.68 & 22.96 & 83.20 & 64.60 & 27.71 & 48.68 \\
Qwen2.5-VL-7B  \cite{bai2025qwen2}      & \underline{75.29} & 54.78 & 14.30 & 33.96 & 89.40 & 54.20 &  9.17 & 53.66 \\
Qwen2.5-VL-32B \cite{bai2025qwen2}      & 75.25 & \underline{63.57} & \underline{19.24} & \textbf{49.24} & 88.70 & \underline{70.00} & \underline{31.87} & 55.46 \\
Qwen2.5-VL-72B  \cite{bai2025qwen2}     & \textbf{82.28} & \textbf{66.68} & \textbf{21.01} & 40.09 & 88.10 & 70.30 & 26.94 & \textbf{61.41} \\
MIMO-VL-SFT \cite{coreteam2025mimovltechnicalreport}         & 70.96 & 45.44 & 14.10 & 44.13 & 84.50 & 58.50 & 22.22 & 46.63 \\
MIMO-VL-RL \cite{coreteam2025mimovltechnicalreport}          & 73.56 & 51.67 & 15.55 & \underline{46.78} & \underline{90.00} & 60.80 & 17.22 & 49.67 \\
\midrule
\rowcolor{gray!20}
\multicolumn{9}{l}{\textbf{GUI-specific models}} \\
UI-TARS-2B-SFT \cite{qin2025ui}      & 37.25 & 22.66 &  4.49 & 19.98 & 83.20 & 69.40 & 21.94 & 39.50 \\
UI-TARS-7B-DPO \cite{qin2025ui}       &  3.79 &  1.94 &  1.68 & 12.63 & 85.00 & 75.00 & 28.33 & 30.01 \\
UI-TARS-7B-SFT \cite{qin2025ui}       & 58.87 & 35.00 &  9.99 & 29.03 & 87.40 & \textbf{79.40} & 29.58 & 49.95 \\
UI-TARS-72B-DPO  \cite{qin2025ui}     & 42.52 & 27.61 &  6.66 & 36.57 & 75.70 & 74.40 & 29.21 & 43.91 \\
UI-TARS-72B-SFT  \cite{qin2025ui}     & 70.28 & 45.49 & 10.97 & 41.08 & \textbf{90.10} & \textbf{79.40} & \textbf{33.33} & \underline{56.22} \\
TongUI-3B \cite{zhang2025tongui}            & 34.66 & 16.19 &  2.63 & 18.27 & 82.20 & 54.70 & 10.83 & 25.29 \\
TongUI-7B \cite{zhang2025tongui}            & 47.31 & 23.62 &  3.81 & 27.49 & 84.80 & 47.70 & 11.39 & 29.99 \\
OS-Atlas-base-7B \cite{wuatlas}       & 40.54 & 31.78 &  6.40 & 19.52 & 81.30 & 50.80 & 23.75 & 28.80 \\
\bottomrule
\end{tabular}}
\label{all}
\end{table*}

\paragraph{Data Annotation.}  For each subtask, candidate instances are selected from single-image and trajectory data and annotated under a unified protocol, comprising four stages: question construction, answer labeling, distractor option generation, and multi-stage quality review. Deterministic labels (e.g., app identification, OCR) are derived from metadata, while interpretive answers are drafted with GPT-4o and verified by humans. Distractors are crafted to remain close to, yet clearly separable from, the correct answer. For grounding, annotators create intuitive target descriptions and label both bounding boxes and click points. For offline agent tasks, the consistency between instructions and trajectories is checked, while online agent tasks are verified for task success and reachability.

\subsection{Quality Control}


To ensure high-quality test items, we adopt a three-stage quality control pipeline, while details for each task can be seen in Appendix A.3:
\paragraph{Stage 1: Manual Cross-Checking.}  Independent annotators review each other’s labels to reduce individual bias, resolve ambiguities, and improve overall consistency.
\paragraph{Stage 2: Automated Quality Inspection.} We first run strong multimodal models over the pool to surface weak spots and frequent errors. Based on their feedback, we strengthen underpowered distractors, replace overly direct or too-easy grounding instructions, and calibrate phrasing to control ambiguity. Relying only on strong models can “guess” correct answers and hide flaws, so we re-check items that passed with a smaller model (2B). Items that also pass stably are kept; otherwise they are revised or removed. 
\paragraph{Stage 3: Manual Evaluation.} We randomly sample 20\% of the items and have annotators complete them under the evaluation protocol to validate item soundness and establish human baselines for each task. This process results in the final benchmark shown in Fig.~\ref{static}.

\section{Experiment}






To comprehensively evaluate the capabilities of current multimodal large language models and agents on Chinese mobile GUI tasks, we select 20 representative models and systems, spanning 47 configurations. The suite covers general-purpose multimodal models, GUI-tuned domain models, and multi-agent systems.

\begin{itemize}
    \item \textbf{General multimodal models.} We include the closed-source baselines GPT-4o and GPT-4o-mini, which perform strongly on multilingual and multimodal tasks. Among open-source models, the Qwen2.5-VL family \cite{bai2025qwen2}, Intern2.5-VL \cite{chen2024expanding}, and Intern3-VL \cite{zhu2025internvl3} demonstrate strong perception, grounding, and transfer abilities across diverse tasks. We additionally include the MIMO-VL series \cite{coreteam2025mimovltechnicalreport}, noted for its solid reasoning performance.
  \item \textbf{GUI-specific models.} The suite covers UI-TARS \cite{qin2025ui}, AgentCPM \cite{zhang2025agentcpm}, UGround \cite{qian2025uground}, OS-Atlas \cite{wuatlas}, TongUI \cite{zhang2025tongui}, ShowUI \cite{lin2025showui}, AGUVIS \cite{xuaguvis}, SeeClick \cite{cheng2024seeclick}, and CogAgent \cite{hong2024cogagent}. We also include r1–style models such as GUI-R1 \cite{luo2025gui}, UI-R1 \cite{lu2025ui}, and InfiGUI-R1 \cite{liu2025infigui} to observe how policy optimization affects practical control ability.
  \item \textbf{Multi-agent systems.} We select Mobile-Agent v1 \cite{wang2024mobile} and v2~\cite{wang2024mobilev2} to assess the performance on real mobile devices.
\end{itemize}

\subsection{Main Results}

In this section, we report the overall performance across all capability dimensions, as well as the best results for each subtask. Since some models only support a subset of application tasks or lack general dialog capabilities, they cannot be directly compared by aggregated scores in Tab.~\ref{all}; their results across all subtasks can instead be found in Appendix A.4. Tab.~\ref{all} presents the performance of representative closed-source models, open-source general-purpose models, and domain-fine-tuned models on GUI-CEval, reflecting the current capability level of Chinese mobile GUI agents in real-world scenarios.

On the GUI-CEval benchmark, the evaluated models exhibit a clear performance difference. Qwen2.5-VL-72B achieves the highest overall average of 61.41\%, followed by the domain-specialized UI-TARS-72B-SFT (56.22\%) and Qwen2.5-VL-32B (55.46\%). In contrast, the closed-source models GPT-4o and GPT-4o-mini reach only 27.69\% and 20.93\%, respectively, falling significantly behind open-source models of similar scale. When examined by capability dimension, perception-related scores such as Perception and Grounding remain high (peaking at 82.28\% and 90.10\%, respectively), while reasoning and correction-related metrics (Reflection and Evaluation) lag far behind (maximums of 21.01\% and 49.24\%). For Online Agent—the most challenging task involving real interactions—the best success rate is 33.33\% (UI-TARS-72B-SFT), with over half of the models below 20\%. Model scaling brings consistent but diminishing returns: within the Qwen family, increasing parameters from 3B to 72B yields average improvements of 5.0\%, 1.8\%, and 5.9\%, yet online success rises only by 4.9\%. Overall, visual understanding has largely matured, whereas robust planning and stable execution after perception remain key bottlenecks. Moreover, small- to medium-scale models with task-specific fine-tuning achieve approximately 70\% accuracy in offline evaluations, underscoring the crucial role of data quality and training strategies. Based on these empirical observations, our main conclusions are as follows.

\begin{enumerate}
    \item \textbf{GUI-CEval is highly challenging.} Even the strongest models show limited overall and online performance, indicating that current systems remain far from real-world deployability.
    
    \item \textbf{Strong perception does not guarantee reliable execution.} While perception and grounding abilities are relatively mature, reasoning, planning, and stable execution remain major bottlenecks for GUI agents.
    
    \item \textbf{Scaling helps but with diminishing returns.} Larger models improve performance, but gains shrink with size. Meanwhile, small- and medium-scale models with targeted fine-tuning can still achieve competitive results.
    
    \item \textbf{SFT builds core ability, RL improves generalization.} Supervised fine-tuning forms the foundation, while moderate reinforcement or preference optimization enhances real-world stability and recovery behavior.
    
    \item \textbf{End-to-end and multi-agent architectures each offer advantages.} Hybrid designs that fuse the strengths of end-to-end and multi-agent architectures may offer the most practical balance between capability and efficiency.
\end{enumerate}

\begin{figure}[t]
    \centering
    \includegraphics[width=\linewidth]{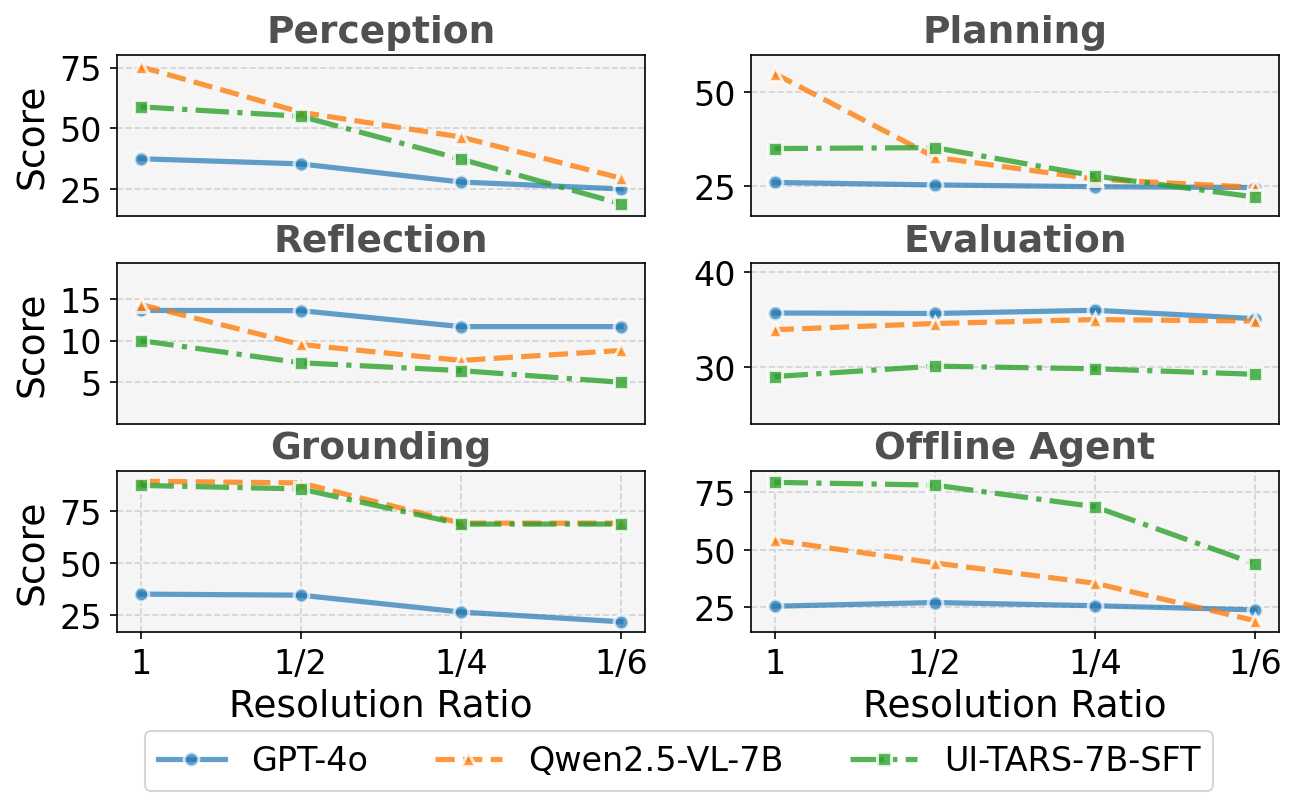}
    \caption{Performance trends of representative models across six capability dimensions under varying input resolutions.}
    \label{resolution}
\end{figure}

\subsection{Fine-grained Analysis}
In this section, we present more fine-grained analysis based on the evaluation results.


\paragraph{Impact of the resolution.} Existing works prefer training and evaluation with high-quality images, while in practical deployment, low resolution is the norm, as on-device inference is constrained by limited memory and bandwidth, and screenshots must often be compressed for transmission. After scaling the width and height of original screenshots from phones and tablets to one half, one quarter, and one sixth respectively, we observe from Fig. \ref{resolution} that \textbf{Perception} and \textbf{Planning} tasks are the most sensitive to resolution: their accuracy drops sharply when the pixel count is halved, and degrades drastically at one sixth. The performance of \textbf{Offline Agent} tasks is similarly affected, due to the compounded loss of information. \textbf{Grounding} is moderately impacted—fine-grained localization begins to deteriorate noticeably beyond one quarter. In contrast, \textbf{Reflection} and \textbf{Evaluation} are the least sensitive to resolution, as they rely more on upstream reasoning chains and historical context rather than fine-grained visual cues. Overall, one-half scaling still preserves acceptable performance, while further compression to one sixth leads to a pronounced collapse in performance across multiple task categories.

\begin{figure}[t]
    \centering
    \includegraphics[scale=0.1]{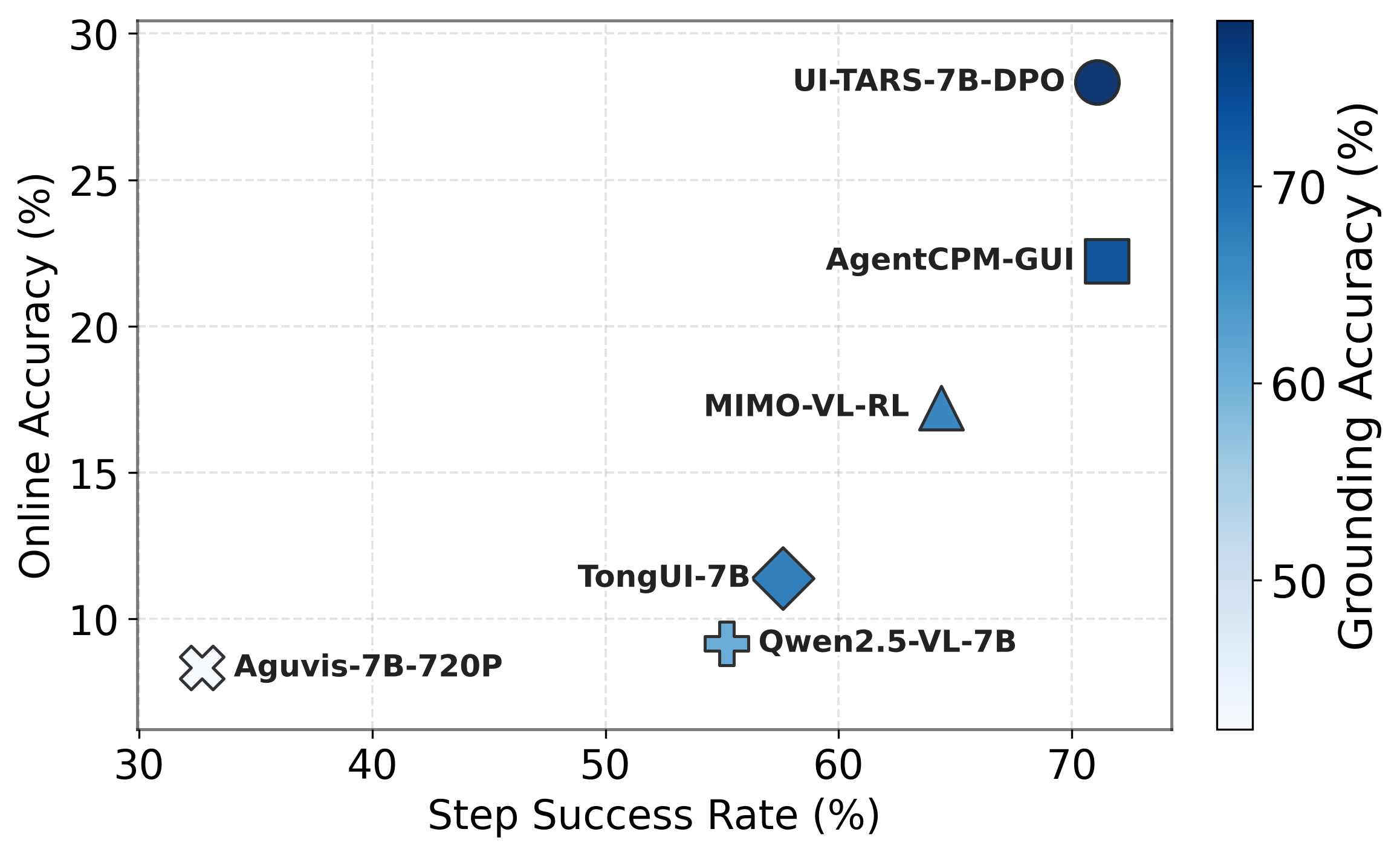}
    \caption{Relationship between step success rate and online task accuracy for representative GUI agents in online environments, with color indicating grounding accuracy.}
    \label{goo}
\end{figure}


\paragraph{Relationships among the three application-level task metrics.}
To analyze the relationships among the three GUI-CEval application tasks systematically, we collected offline execution traces for the same instructions used in Online Agent task and computed step success rate (SSR). To avoid perception bias from mixed action types, Grounding accuracy is measured only on samples where both annotation and prediction are clicks, yielding a cleaner grounding metric. Using this protocol, we compare seven representative systems, as shown in Fig.~\ref{goo}. The results reveal two main trends. First, SSR and Online accuracy are monotonically correlated but nonlinear: beyond the 50–60\% SSR range, improvements in SSR yield diminishing returns in Online due to long-horizon instability, rollback limitations, and system perturbations. Second, Grounding positively influences both SSR and Online, but not in a one-to-one manner. UI-TARS-7B-DPO leads in Grounding and SSR and achieves the highest Online, consistent with the progression from accurate localization to stable single steps to stronger end-to-end success. In contrast, TongUI-7B and Qwen2.5-VL-7B exhibit similar SSR but different Online performance, showing that online success depends heavily on factors beyond perception and single-step reliability. In summary, SSR provides a cost-effective and directionally reliable proxy for early-stage model evaluation, while Online remains the definitive indicator of real deployment capability.

\begin{figure}[!t]
    \centering
    \includegraphics[scale=0.1]{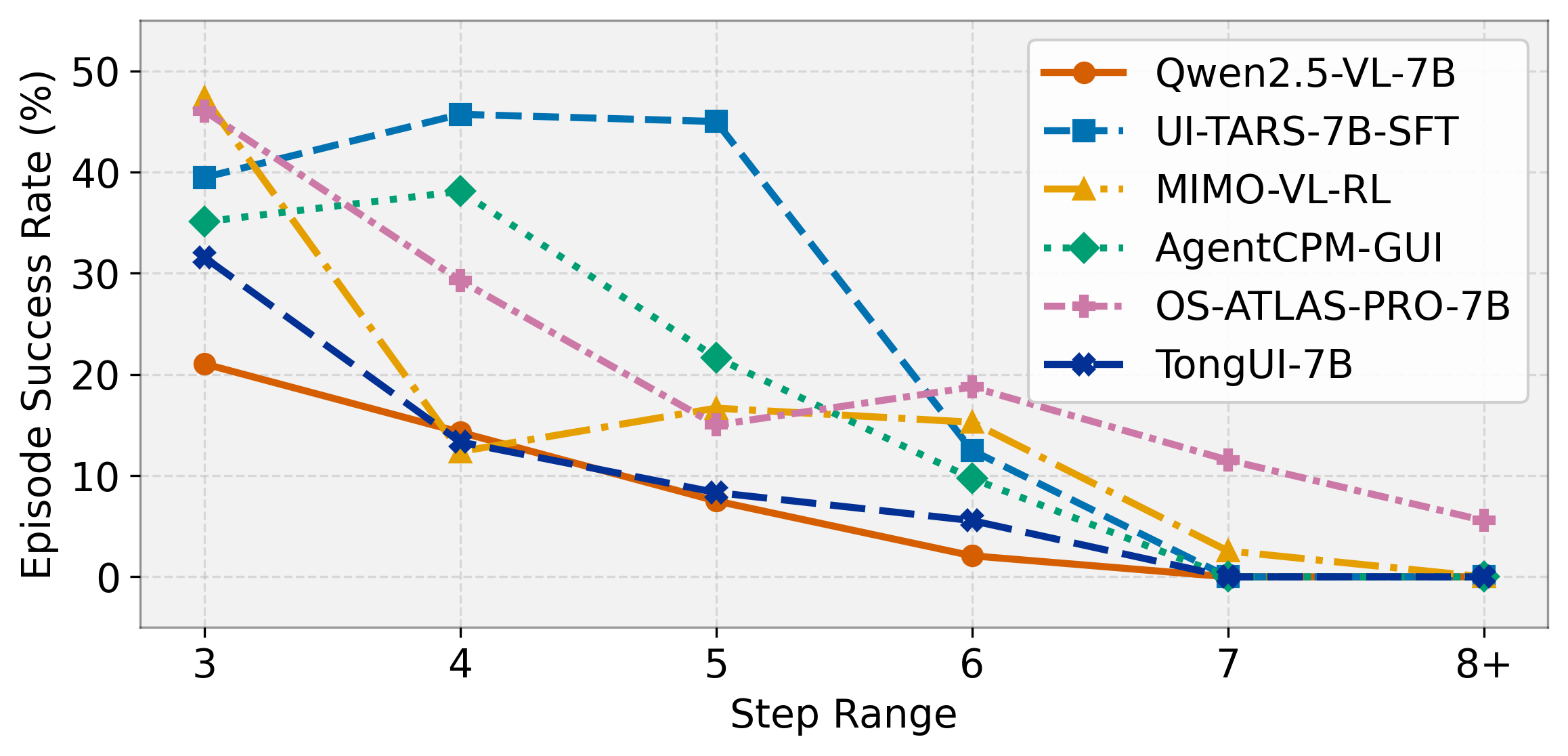}
    \caption{Episode success rates of representative GUI agents across different step ranges, illustrating the sharp performance decline as task length increases.}
    \label{esr}
\end{figure}


\paragraph{Effect of task length on success rate.}
Episode length is the most direct and quantifiable variable of online task difficulty. As shown in Fig.~\ref{esr}, when a task requires only 3 steps, most models cluster around an online success rate of 30–58\%; when the step count increases to 4--5, success rates rapidly drop into the 8–46\% range; once the count exceeds 6 steps, the vast majority of models fall to 2–19\%; at 7 steps and above, performance is almost an “across-the-board collapse”: most approach 0, with only a few remaining in single digits. This monotonic and “cliff-like” decline with increasing steps indicates that the length of the execution chain directly governs error accumulation and state drift, making it an effective proxy for online task difficulty. Other factors such as application domain or page style also influence performance, but their dimensions are numerous and distributions sparse, making unified quantification difficult in the current evaluation. Therefore, future model training should prioritize long-horizon, multi-step tasks, improving robustness at high step counts via process supervision, fault-tolerant strategies, and multi-path data.


\begin{figure}[!t]
    \centering
    \includegraphics[scale=0.08]{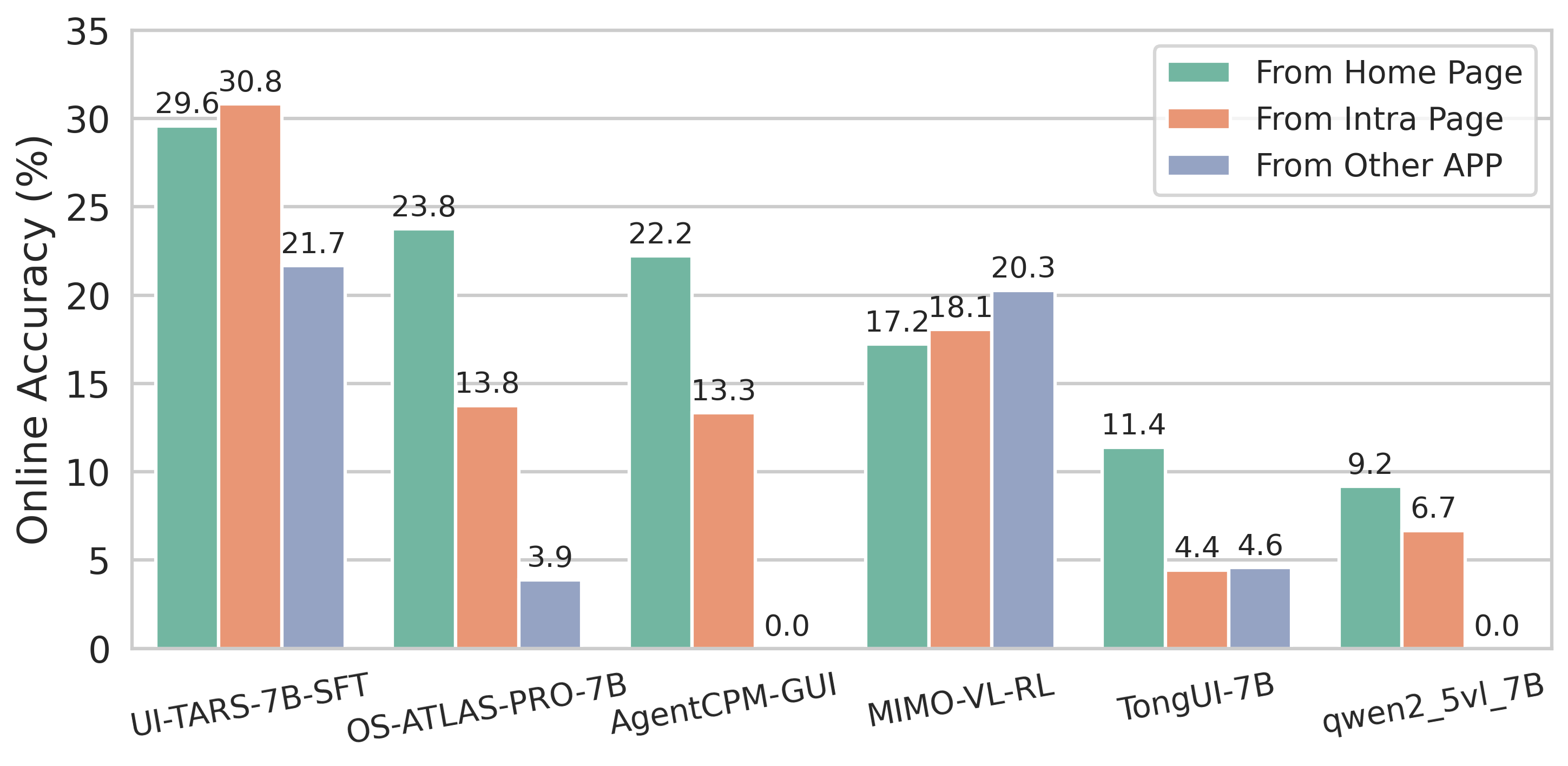}
    \caption{Comparison of online success rates under different initialization conditions (Home, Intra, Other) for representative GUI agents.}
    \label{online}
\end{figure}

\paragraph{Impact of the initial page on online testing.}
Fig.~\ref{online} compares the effect of three initial conditions on the success rates of online tasks. Here, \textit{Home} denotes the home screen of the device, Intra denotes the home page \textit{of} the app, and \textit{Other} denotes the home page of the Xiaomi Auto app uniformly. Compared with the common practice of launching from Home, we observe a significant performance drop once the initial state deviates from the home screen. For example, Qwen2.5-VL-3B decreases from 27.7\% to 12.5\% and further drops to 0 in the Other condition; OS-ATLAS-PRO-7B similarly declines from 23.8\% to 3.9\%. Even the leading model UI-TARS-7B-SFT falls to 21.7\% under the Other setting. This suggests that current GUI agents struggle with state awareness and policy transfer when confronted with diverse and uncontrolled initial pages on real devices.
Interestingly, MIMO-VL-RL is a notable counterexample, achieving a higher success rate of 20.3\% in the Other condition. We speculate that this gain stems from its use of large-scale long-horizon Chinese GUI trajectories and strengthened reasoning. In general, the results show that a single “standard launch flow’’ cannot represent real distributions of user paths. Future models should incorporate multi-start, multi-path, and multi-interruption scenarios into training and evaluation, and leverage long-horizon data and process supervision to enhance robustness and generalization in online environments.

\section{Conclusion}

In this study, we introduce GUI-CEval, a comprehensive benchmark designed to fill the gap in evaluating GUI agents under real Chinese mobile environments. Built upon 201 mainstream applications across four device types, GUI-CEval adopts a two-tier structure covering both fundamental and application-level capabilities, offering a unified and diagnostically meaningful view of the full workflow of GUI agents—from perception and planning to execution, reflection, and evaluation. The human-reviewed data pipeline ensures realistic task contexts and reliable annotations, mitigating language bias, platform heterogeneity, and the limitations of automatic data construction. Extensive evaluation of 20 representative models and systems shows that, despite rapid advances in MLLMs, current approaches still struggle with generalization and stability in real-world scenarios, with notable weaknesses in reflection and evaluation. GUI-CEval thus provides fine-grained capability analyses, actionable feedback for system improvement, and a solid foundation for future work toward more robust and deployable Chinese mobile GUI agents.
{
    \small
    \bibliographystyle{ieeenat_fullname}
    \bibliography{main}
}


\end{document}